\documentclass[letterpaper, 10 pt, conference]{ieeeconf}  

\IEEEoverridecommandlockouts                              

\usepackage{multirow}
\usepackage{graphicx}
\usepackage{amsfonts}
\usepackage{hyperref}

\title{\LARGE \bf
RGB-based Category-level Object Pose Estimation via Decoupled Metric Scale Recovery
}

\author{Jiaxin Wei$^{1}$, Xibin Song$^{2}$, Weizhe Liu$^{2}$, Laurent Kneip$^{1}$, Hongdong Li$^{2,3}$ and Pan Ji$^{2}$
\thanks{$^{1}$Mobile Perception Lab, School of Information Science and Technology, ShanghaiTech University;}%
\thanks{$^{2}$Tencent;}%
\thanks{$^{3}$Australian National University.}%
}

\begin{document}

\maketitle
\thispagestyle{empty}
\pagestyle{empty}

\begin{abstract}

While showing promising results, recent RGB-D camera-based category-level object pose estimation methods have restricted applications due to the heavy reliance on depth sensors. RGB-only methods provide an alternative to this problem yet suffer from inherent scale ambiguity stemming from monocular observations. In this paper, we propose a novel pipeline that decouples the 6D pose and size estimation to mitigate the influence of imperfect scales on rigid transformations. Specifically, we leverage a pre-trained monocular estimator to extract local geometric information, mainly facilitating the search for inlier 2D-3D correspondence. Meanwhile, a separate branch is designed to directly recover the metric scale of the object based on category-level statistics. Finally, we advocate using the RANSAC-P$n$P algorithm to robustly solve for 6D object pose. Extensive experiments have been conducted on both synthetic and real datasets, demonstrating the superior performance of our method over previous state-of-the-art RGB-based approaches, especially in terms of rotation accuracy. Code: \url{https://github.com/goldoak/DMSR}.

\end{abstract}


\section{INTRODUCTION}

Accurately estimating the position and the orientation of an object in 3D space is critical for perceiving surrounding environments, thus has broad applications in computer vision~\cite{tan20176d, su2019deep} and robotics communities~\cite{du2021vision, wang2020feature}. Previous efforts in 6D object pose estimation have largely focused on the instance-level setting~\cite{xu2022rnnpose, iwase2021repose} where a corresponding CAD model is given for each object of interest. In such cases, object pose estimation can be simplified to correspondence matching between 3D models and observations. However, the requirement of prior CAD models for each entity is not only hard to achieve in real-world scenarios, but also makes it difficult to scale well to complicated scenes.

To relax the need for instance-level 3D models, category-level object pose estimation~\cite{lin2022category, chen2021sgpa, lin2022sar} has been intensively investigated recently. A special kind of coordinates proposed by~\cite{nocs} are generated from Normalized Object Coordinate Spaces (NOCS) to align different instances into a normalized canonical space in a category-wise manner. The prediction of rotation and translation is thus performed within the same category as the target object. This line of research mainly deals with the intra-class shape variation~\cite{di2022gpv, lin2021dualposenet, chen2021fs}. Once a set of inlier 3D-3D correspondences is found, accurate pose and size can be estimated using similarity transformation. Promising results have been obtained by these methods, yet heavily relying on depth observations to achieve accurate pose and size estimation limits their broader applications.

Therefore, RGB-based category-level object pose estimation~\cite{chen2020synthesis, fan2022depth} has emerged in this direction. They can be a viable alternative when depth measurements are not available, such as on VR headsets and mobile devices. However, directly predicting object pose from a single RGB image remains a challenging task nowadays as it is a highly ill-posed problem, suffering from inherent scale ambiguity.~\cite{lee2021shape} and~\cite{fan2022depth} propose first to reconstruct the absolute depth map of objects, thereby transferring the problem to the familiar RGBD setting. Then again, 3D-3D correspondences are used to solve for the object pose and size at the same time. Though intuitive, predictions of metric depth maps are often inaccurate, leading to unsatisfactory results. It also makes these methods hard to generalize to different scenes.

Unlike previous RGB-based methods expecting to implicitly learn the metric scale through absolute depth prediction, we instead choose to decouple object pose and size estimation for RGB-based category-level pose estimation. The motivation behind this choice is obvious: we want to explicitly cut off the propagation of errors from metric scale to 6D pose estimation, especially for rotation. To this end, we introduce a novel pipeline for RGB-based category-level object pose and size estimation, consisting of feature extraction, 2D-3D correspondence learning, metric scale recovery, and pose estimation. Specifically, the (relative) depth and normal are taken as additional input to exploit instance-level geometric information. They are generated by readily available models pre-trained on a large dataset, Omnidata~\cite{eftekhar2021omnidata} for each RGB image. Inspired by~\cite{chen2021sgpa} and~\cite{tian2020shape-prior}, we employ a small transformer to extract distinctive features for subsequent correspondence learning. Meanwhile, we design a simple but effective scale prediction branch to recover the metric scale of the target object by leveraging category-level statistics. Finally, we compute the 6D object pose using a traditional P$n$P solver inside a RANSAC loop to deal with potential outliers. Our main contributions can be summarized as follows:
\begin{enumerate}
    \item[(a)] We propose the decoupled framework for RGB-based category-level object pose estimation, where 6D object pose and size are computed separately to reduce the negative influence of imperfect scale predictions on rigid transformations.
    
    \item[(b)] The metric scale of the target object is directly recovered from the network on the basis of category-level priors and is subsequently used for robust pose estimation in a RANSAC-P$n$P algorithm.
    
    \item[(c)] We extend the CAMERA25 and REAL275 datasets to include depth and normal predictions from pre-trained models. Our method is simple but effective, achieving significant improvements over previous RGB-based category-level pose estimation methods, particularly in terms of rotation accuracy.
\end{enumerate}

\begin{figure*}[htbp]
\centering
\includegraphics[width=\linewidth]{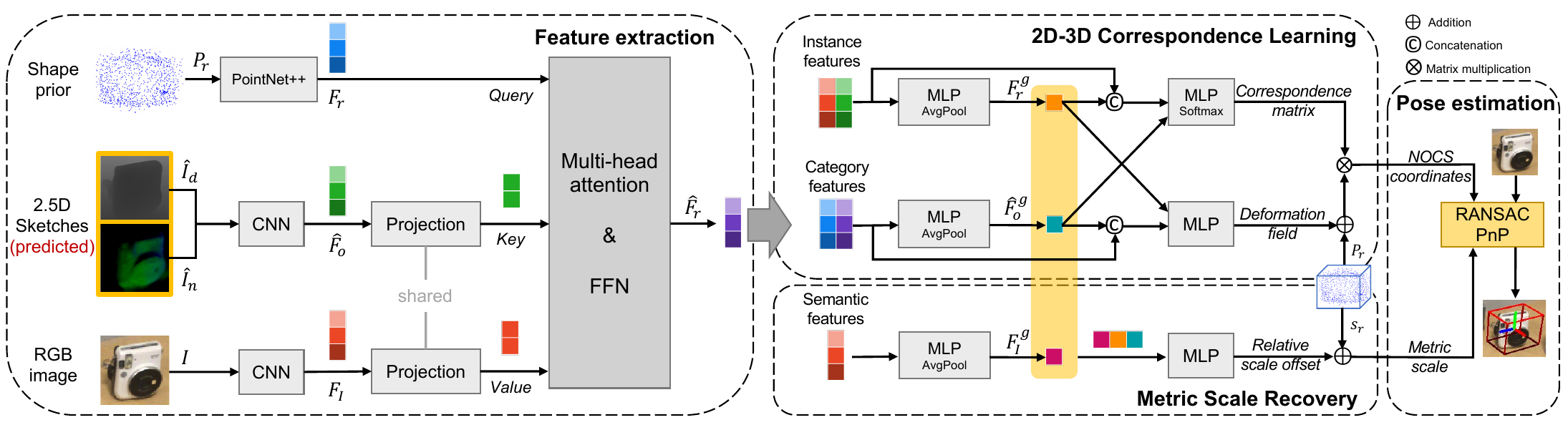}
\caption{Illustration of our pipeline for RGB-based category-level object pose estimation.}
\label{Fig:pipeline}
\end{figure*}

\section{RELATED WORK}

Image-based object pose estimation has a long-standing history and was first addressed at the instance-level and based on exact object shape priors. The category-level case---as addressed in this work---has only been studied much later. A good overview of the field including but not limited to category-level object pose estimation can be found in recent survey papers such as the work of~\cite{bruns2023rgbdbased}. From a high-level point of view, methods can be classified depending on whether or not they make use of depth readings during inference time.

\subsection{RGBD-based Category-level Object Pose Estimation}

A seminal contribution towards category-level pose estimation is presented by~\cite{nocs}, who introduce the Normalized Object Coordinate Space (NOCS). In this space, all objects within the category exhibit the same alignment as defined by object-specific directions~\cite{nocs} introduce a framework that relies on Mask R-CNN~\cite{he2017maskrcnn} to detect objects and an additional head to predict the projection of these coordinates into the image plane (denoted the NOCS-map). Using a scale-aware implementation of Procrustes alignment~\cite{umeyama1991umeyama}, the object pose can be found.~\cite{chen2020learning} propose an alternative approach in which both image and depth readings are fed to a variational auto-encoder to directly regress the object position and orientation. Later on,~\cite{tian2020shape-prior} then introduces the idea of Shape Prior Deformation (SPD), an approach in which a canonical point set is fed to a deformation network. The deformed canonical point set is again aligned with the input depth readings (or a derived representation) for object pose estimation.~\cite{wang2021cascaded} extend the architecture by a recurrent network in which the canonical point set is deformed iteratively.~\cite{chen2021sgpa} finally introduce SGPA, a transformer-based architecture to more effectively adjust the canonical point set. Though requiring RGB-D input, SGPA serves as a strong inspiration for the method proposed in this paper. ~\cite{lin2022selfsupervised} make use of consistency-based losses in order to improve learning by self-supervision. An alternative line of works that also in return produce category-level object pose, but is more computationally demanding, is given by analysis-by-synthesis methods that iteratively deform and align an implicit shape model until convergence.

\subsection{RGB-based Category-level Object Pose Estimation}

One of the first analysis-by-synthesis approaches was indeed proposed for the pure RGB case.~\cite{chen2020synthesis} notably make use of an implicit neural representation for iterative shape synthesis and object pose estimation. A disadvantage of the method however consists of its inability to recover correctly scaled object pose results, as the method only evaluates alignment in the image.~\cite{manhardt2020cps} propose a straightforward regression network for object pose and shape trained on synthetic data. While they also propose a depth-supported refinement strategy to close the synthetic-to-real gap, the majority of their results focus on the synthetic case. Another well-performing method is introduced by~\cite{lee2021shape}, who propose a two-branch network to predict a metrically scaled object mesh and a NOCS-map. After rendering a depth map from the reprojected mesh, a concluding geometric alignment step returns the object pose. Finally,~\cite{fan2022depth} propose OLD-Net, a variation that directly predicts object-level depths by incorporating global position hints and shape priors.

In a word, the focus of existing attempts to recover metrically scaled results consists of recovering an absolute depth map from the input measurements. It is well-understood that this is a challenging problem. However, in this work, we propose a novel RGB camera-based category-level pose estimation framework that employs a decoupled scale estimation branch in order to scale a canonical object point set prior back to the metric space, without interfering with 6D object pose estimation. While we still depend on a relative depth prediction, depth uncertainties are further prevented from entering the final pose estimation by using perspective rather than generalized Procrustes alignment.

\section{APPROACH}

\subsection{Overview}

Given a single RGB image, our goal is to estimate the 6DoF object pose (3D rotation and 3D translation) with respect to the camera coordinate frame, as well as the metric scale of the object. As shown in Figure \ref{Fig:pipeline}, the pipeline mainly consists of four parts: feature extraction, 2D-3D correspondence learning, metric scale recovery, and pose estimation. We first pre-process each image using models pre-trained on the large Omnidata to generate local geometric cues as described in Section \ref{Sec:dpt} and then feed them into the network together with category-level shape priors and RGB images to further extract higher-level features for 2D-3D correspondence learning (Section \ref{Sec:corr}). A separate branch for metric scale recovery is described in Section \ref{Sec:scale}, which is essential for the final pose estimation with P$n$P algorithm (Section \ref{Sec:pnp}). 

\subsection{Leveraging Geometric Information} \label{Sec:dpt}

The network is supposed to find as many inlier 2D-3D correspondences as possible. However, there is no depth measurement in the RGB-only setting. To compensate for the missing instance-level geometric information, we turn to use an off-the-shelf monocular estimator to predict 2.5D sketches~\cite{Marr1982}, e.g. depth and normal maps. In particular,~\cite{eftekhar2021omnidata} has built an Omnidata pipeline to train strong models on a large-scale dataset, which achieves remarkable performance regarding both accuracy and generalization ability. 

Therefore, in a preprocessing step, we feed each image patch $I \in \mathbb{R}^{H\times W \times 3}$ through a depth estimation model $f_d$ and surface normal estimation model $f_n$ to obtain $\hat{I_d} = f_d(I), \hat{I_n} = f_n(I)$ where $\hat{I_d} \in \mathbb{R}^{H\times W \times 1}$ and $\hat{I_n} \in \mathbb{R}^{H\times W \times 3}$ are depth and normal maps, respectively. It is worth noting that $\hat{I_d}$ is a relative depth cue, merely depicting the geometric relationship among object pixels. Without knowing the metric scale, it is meaningless to compute 3D point cloud back-projected from $\hat{I_d}$. Also, the inaccurate depth predictions on the edges of the object further prohibit the direct application of point clouds. 

\subsection{2D-3D Correspondence Learning} \label{Sec:corr}

Our network takes cropped image patch $I \in \mathbb{R}^{H\times W \times 3}$, category-level shape prior $P_r \in \mathbb{R}^{N\times 3}$, and predicted 2.5D sketches $\hat{I_d} \in \mathbb{R}^{H\times W \times 1}$ and $\hat{I_n} \in \mathbb{R}^{H\times W \times 3}$ as inputs. Note that the corresponding class label and object mask can be obtained from an off-the-shelf instance segmentation network, such as Mask R-CNN~\cite{he2017maskrcnn}. These inputs are then fed into a feature extraction module to obtain the semantic feature $F_I \in \mathbb{R}^{M\times d}$, geometric features $F_r \in \mathbb{R}^{N\times d}$ and $\hat{F_o} \in \mathbb{R}^{M\times d}$, respectively. The $M$ pixel features in $F_I$ are randomly selected to reduce the number of correspondences. In order to mitigate the interference of noise, we apply channel-wise concatenation on $\hat{I_d}$ and $\hat{I_n}$, and then use a fully convolutional network $g$, following an encoder-decoder architecture, to generate instance-specific geometric features $\hat{F_o} = g([\hat{I_d}| \hat{I_n}])$ where $[\cdot|\cdot]$ denotes channel-wise concatenation. Intuitively, we expect $\hat{F_o}$ can benefit from the integration of these two kinds of geometric cues. This argument is later supported by our experiments in Section \ref{Sec:exp_ablation}.

Following~\cite{tian2020shape-prior} and~\cite{chen2021sgpa}, we take advantage of category-level prior models on which instance-specific deformation is performed to reconstruct shape details. Hence, the resulting NOCS coordinates will represent the desired shape of the input object. To better guide the deformation of the shape prior, $F_r, \hat{F_o}, F_I$ are considered as the query, key and value of a transformer such that the semantic information in $F_I$ can be adaptively propagated to prior $F_r$, according to the structural similarity between $F_r$ and $\hat{F_o}$. Here $F_r$ and $\hat{F_o}$ are projected to a lower dimension such that only object keypoints are taken into account to improve computation efficiency. The adapted prior feature $\hat{F}_r$ is then concatenated with $F_r$ to further predict the deformation field $D$, thereby resulting in the final reconstructed model $P_{nocs} = P_r + D$. Meanwhile, the matching network outputs the correspondence matrix $C \in \mathbb{R}^{M\times N}$ using the concatenation of $F_I$ and $\hat{F_o}$. 

\subsection{Metric Scale Recovery} \label{Sec:scale}

The point cloud back-projected from the observed depth map is the key to reliable object pose and size estimation for those RGBD-based methods, as the underlying metric scale of depth measurements can significantly ease the difficulty of locating the object in 3D space. However, we make use of the scale-agnostic depth prediction $\hat{I_d}$ and try to solve for the metric scale in a separate way.

Direct regression of metric scales seems to be the simplest solution, but the metric scales are not fixed into a stable range, making the network hard to train. Instead, we take advantage of category-level statistics and reduce the problem to an offset regression task. Similar to the generation of category-level shape prior $P_r$ (i.e. mean shape), we also compute the mean scale $s_r \in \mathbb{R}$ for each category and set it as an anchor for scale prediction to avoid large deviations from desired results. Then, we pass the concatenation of extracted global features $F_r^g, \hat{F_o^g}, F_I^g \in \mathbb{R}^m$ from $F_r, \hat{F_o}, F_I$ through a multi-layer perceptron (MLP) to recover the relative scale offset
\begin{equation}
    \Delta s = \mathrm{MLP}([F_r^g| \hat{F_o^g}| F_I^g])\quad s.t. \quad \hat{s} = s_r + s_r \times \Delta s.
\end{equation}
Here, we employ the $\mathcal{L}1$ loss to supervise the training of our scale recovery branch
\begin{equation}
    L_{scale} = \| \Delta s_{gt} - \Delta s\|, \Delta s_{gt} = \frac{s_{gt} - s_r}{s_r}.
\end{equation}
In this way, we manage to obtain the metric scale for subsequent pose estimation.

\begin{figure*}[htbp]
\centering
\includegraphics[width=\linewidth]{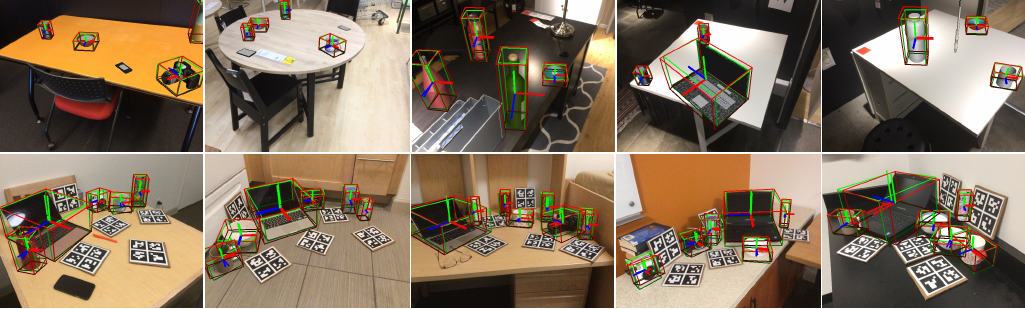}
\caption{Qualitative results of our method on CAMERA25 (top) and REAL275 (bottom). The predictions and ground truths are shown by red and green bounding boxes, respectively.}
\label{Fig:vis}
\end{figure*}

\subsection{Perspective-n-Point for Pose Estimation} \label{Sec:pnp}

Finally, we use P$n$P algorithm, which takes 2D-3D correspondences as input, to solve for the object pose. Unlike the similarity transformation, P$n$P is incapable of handling NOCS coordinates without scale information. Therefore, we need to scale the NOCS coordinates before feeding them into the solver
\begin{equation}
    \hat{R}, \hat{t} = \mathrm{PnP}(P_I, \hat{s} \cdot CP_{nocs}) = \mathrm{PnP}(P_I, \hat{s} \cdot C(P_r + D)))
\end{equation}
where $P_I \in \mathbb{R}^{M \times 2}$ is the corresponding 2D pixel coordinates on image patch, $\hat{R} \in SO(3)$ is the rotation and $\hat{t} \in \mathbb{R}^{3}$ is the translation.

\subsection{Overall Loss Function} \label{Sec:loss}

Our pipeline mainly comprises two learning objectives, namely the NOCS correspondences and the relative scale offset $\Delta s$. The overall loss function is then the weighted sum of the following two loss terms:
\begin{equation}
    L = \lambda_1 L_{corr} + \lambda_2 L_{scale}. \label{Eq:loss}
\end{equation}
Here, $L_{corr}$ is the same loss used in~\cite{chen2021sgpa}, including the supervision on the deformation field $D$ and the correspondence matrix $C$, as well as the regularization term on the keypoint projection matrix. Please refer to~\cite{chen2021sgpa} for more details of $L_{corr}$.

\subsection{Implementation Details}

We generate depth and normal maps using the latest DPT models~\cite{ranftl2021dpt} pre-trained on Omnidata~\cite{eftekhar2021omnidata} with 3D data augmentations~\cite{kar2022augmentation}. For each image patch, we first resize it to $384 \times 384$, and then feed it into $f_d$ and $f_n$ to obtain $\hat{I_d}$ and $\hat{I_n}$, respectively. The outputs are resized to $192 \times 192$ later on for feature extraction. Specifically, we employ a 4-layer PSPNet~\cite{zhao2017psp} with ResNet18~\cite{he2016resnet} backbone as $g$. As for metric scale recovery, we implement a 3-layer MLP with hidden dimensions 512 and 128 to regress the relative scale offset. To better handle the outliers, we estimate the object pose using a traditional P$n$P solver inside a RANSAC loop. The two weights $\lambda_1, \lambda_2$ in the loss function \ref{Eq:loss} are set to 1.0 and 0.1, respectively. The rest of the architectural details remain the same as~\cite{chen2021sgpa}. All experiments are conducted on two NVIDIA RTX 2080Ti GPUs.

\section{Experiments}

\subsection{Datasets}

We conduct extensive experiments on two benchmark datasets released by~\cite{nocs} for the category-level object pose estimation task. The CAMERA dataset is synthesized by placing CAD models on top of real backgrounds while the REAL dataset is directly collected in the real world. These two datasets contain 6 categories (i.e. bottle, bowl, camera, can, laptop and mug), and we can generate a mean shape as prior for each category following~\cite{tian2020shape-prior}. Note that CAMERA25 and REAL275 are two subsets for testing, consisting of 25k synthetic images and 2750 real images, respectively. We also extend the NOCS datasets by predicting depth and normal maps for each image patch cropped from the original $640 \times 480$ RGB image.

\subsection{Metrics}

We adopt the widely used evaluation metrics in category-level object pose estimation for a fair comparison. The first one is 3D Intersection-Over-Union (IoU) under different thresholds. It measures the IoU between two 3D bounding boxes transformed by estimated pose and ground truth, mainly reflecting the fidelity of object size and translation. For 6D object pose, we directly compute the rotation error in degrees and translation error in centimeters. We report the mean average precision (mAP) across all categories for both metrics in the following tables.

\begin{table*}[htbp]
\centering
\caption{Quantitative comparisons of SOTA RGB-based methods on CAMERA25 and REAL275 datasets.}
\label{Tab:main_exp}
\begin{tabular}{c|ccccc|ccccc}
\hline
\multirow{2}{*}{Method} & \multicolumn{5}{c|}{CAMERA25}                                                 & \multicolumn{5}{c}{REAL275}                                                   \\ \cline{2-11} 
                        & IoU50         & IoU75        & 10cm          & $10^\circ$    & $10^\circ$10cm & IoU50         & IoU75        & 10cm          & $10^\circ$    & $10^\circ$10cm \\ \hline
Synthesis               & -             & -            & -             & -             & -              & -             & -            & 34.0          & 14.2          & 4.8            \\ \hline
MSOS                    & 32.4          & 5.1          & 29.7          & 60.8          & 19.2           & 23.4          & 3.0          & \textbf{39.5} & 29.2          & 9.6            \\ \hline
OLD-Net                 & 32.1          & 5.4          & 30.1          & 74.0          & 23.4           & 25.4          & 1.9          & 38.9          & 37.0          & 9.8            \\ \hline
Ours                    & \textbf{34.6} & \textbf{6.5} & \textbf{32.3} & \textbf{81.4} & \textbf{27.4}  & \textbf{28.3} & \textbf{6.1} & 37.3          & \textbf{59.5} & \textbf{23.6}  \\ \hline
\end{tabular}%
\end{table*}

\begin{table*}[htbp]
\centering
\caption{Validations of our design choice on different input modalities.}
\label{Tab:ablation_input}
\begin{tabular}{ccc|ccccc|ccccc}
\hline
\multicolumn{3}{c|}{Input modality}   & \multicolumn{5}{c|}{CAMERA25}                                                 & \multicolumn{5}{c}{REAL275}                                                   \\ \hline
point cloud & depth      & normal     & IoU50         & IoU75        & 10cm          & $10^\circ$    & $10^\circ$10cm & IoU50         & IoU75        & 10cm          & $10^\circ$    & $10^\circ$10cm \\ \hline
\checkmark  & -          & -          & 33.8          & 6.4          & 32.0          & 76.0          & 25.7           & 23.3          & 6.6          & 33.2          & 43.9          & 19.5           \\ \hline
-           & \checkmark & -          & 34.3          & 6.2          & 31.7          & 79.2          & 26.3           & 23.8          & \textbf{6.9} & 32.4          & 56.8          & 21.5           \\ \hline
-           & \checkmark & \checkmark & \textbf{34.6} & \textbf{6.5} & \textbf{32.3} & \textbf{81.4} & \textbf{27.4}  & \textbf{28.3} & 6.1          & \textbf{37.3} & \textbf{59.5} & \textbf{23.6}  \\ \hline
\end{tabular}%
\end{table*}

\subsection{Comparison against State-of-the-art Methods}

We mainly compare our proposed method with existing state-of-the-art (SOTA) RGB-based methods: Synthesis~\cite{chen2020synthesis}, MSOS~\cite{lee2021shape}, and OLD-Net~\cite{fan2022depth}. As far as we know, these works are the SOTA methods trying to solve category-level pose estimation using a single RGB image. Please note that MSOS can use one pixel depth value per object to boost the performance, we only report their RGB setting for a fair comparison. The quantitative results are shown in Table \ref{Tab:main_exp}. It is obvious to find that our method significantly outperforms previous RGB methods in terms of rotation accuracy (i.e. $10^\circ$), demonstrating the strong ability in correspondence learning. More specifically, our method is 7.4\% and 22.5\% higher than OLD-Net w.r.t. $10^\circ$ metric on CAMERA25 and REAL275 datasets, respectively. For translation and 3D IoU metrics, our method also achieves competitive results, especially for the strictest IoU75 metric on REAL275, where our method is 3.1\% higher than MSOS. This proves the effectiveness of our proposed metric scale recovery. The overall performance of our method under the $10^\circ$10cm metric hence surpasses other approaches by a large margin (4.0\% and 13.8\% higher than OLD-Net on both datasets) due to the improvements in both rotation and translation estimation.

The qualitative results of our method are shown in Figure \ref{Fig:vis}. We visualize the estimated object pose and size in projected 3D bounding boxes (red) and object coordinate axes. Ground truths are also presented in green bounding boxes for reference. It can be seen from the figures that our method is able to generate considerably tight bounding boxes around objects, thereby reflecting its faithfulness in metric scale recovery. Moreover, through the orientation of projected axes, we can discover the prediction consistency across categories.

\begin{figure}
  \centering
  \includegraphics[width=\linewidth]{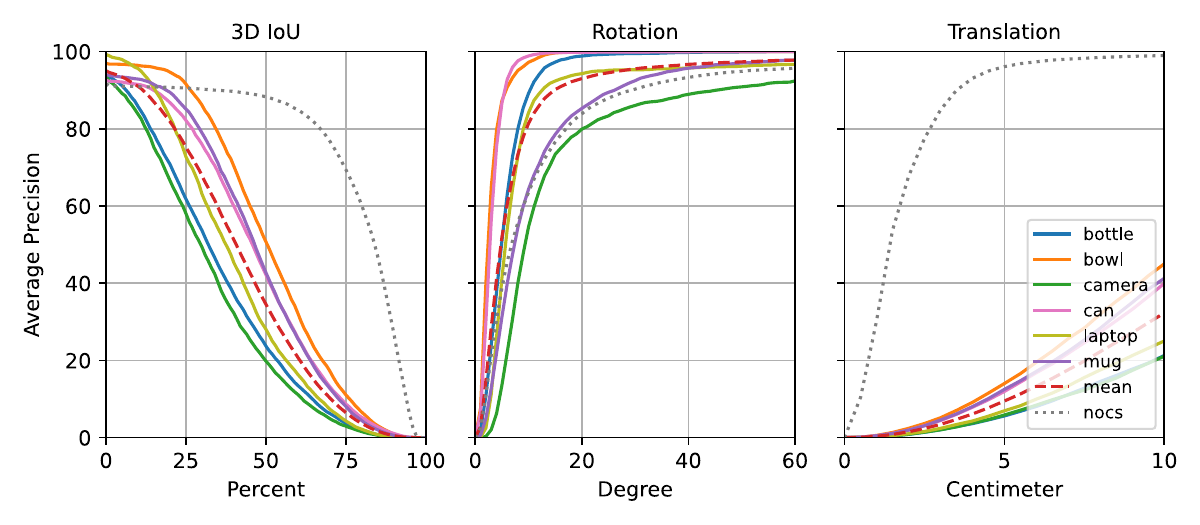}
  \includegraphics[width=\linewidth]{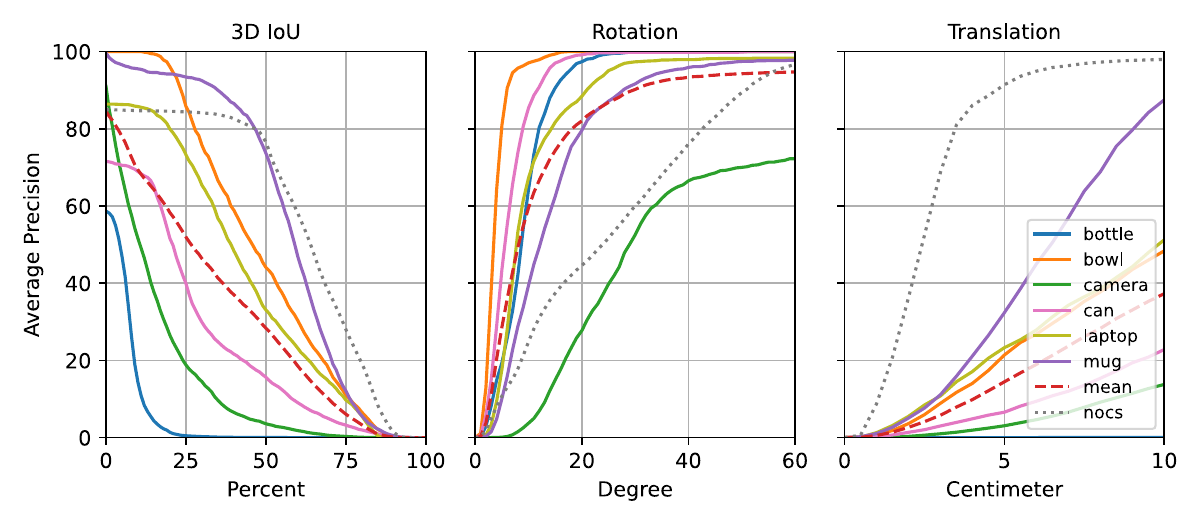}
  \caption{Average precision vs. error thresholds on CAMERA25 (top) and REAL275 (bottom).}
  \label{Fig:mAP}
\end{figure}

We further present a more detailed analysis of average precision (AP) against different error thresholds in Figure \ref{Fig:mAP}. It clearly shows the different performances of each category regarding 3D IoU, rotation error, and translation error. Generally speaking, the performance on CAMERA25 is better than REAL275, partially because of the insufficient training data and cluttered backgrounds in real scenarios. As for evaluation metrics, our method outperforms the baseline NOCS~\cite{nocs}, a RGBD-based method, in terms of rotation error for almost all categories, except for the camera class. By comparing different categories w.r.t. 3D IoU and translation error, we can discover that the performance of the bottle and camera class is often inferior to that of other object categories. This probably results from the large shape variation within these two classes. We will discuss it in detail in Section \ref{Sec:exp_scale}.

\subsection{Ablation Studies} \label{Sec:exp_ablation}

\subsubsection{Different Input Modalities} 

\begin{table*}[htbp]
\centering
\caption{Effects of our proposed scale offset regression branch.}
\label{Tab:ablation_scale}
\begin{tabular}{cc|ccccc|ccccc}
\hline
\multicolumn{2}{c|}{Scale} & \multicolumn{5}{c|}{CAMERA25}                                                 & \multicolumn{5}{c}{REAL275}                                                   \\ \hline
mean scale  & scale offset & IoU50         & IoU75        & 10cm          & $10^\circ$    & $10^\circ$10cm & IoU50         & IoU75        & 10cm          & $10^\circ$    & $10^\circ$10cm \\ \hline
\checkmark  & -            & 23.1          & 3.6          & 21.2          & 81.4          & 18.1           & \textbf{30.1} & 3.1          & \textbf{42.0} & 54.4          & \textbf{26.7}  \\ \hline
\checkmark  & \checkmark   & \textbf{34.6} & \textbf{6.5} & \textbf{32.3} & \textbf{81.4} & \textbf{27.4}  & 28.3          & \textbf{6.1} & 37.3          & \textbf{59.5} & 23.6           \\ \hline
\end{tabular}%
\end{table*}

\begin{figure}
  \centering
  \includegraphics[width=\linewidth]{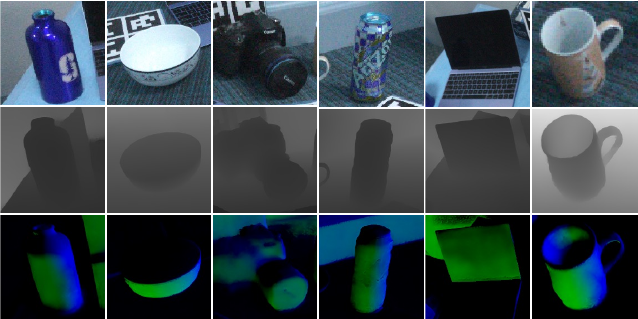}
  \caption{Examples of generated 2.5D sketches. We show the image patch, depth map, and normal map for all 6 categories (i.e. bottle, bowl, camera, can, laptop and mug) on REAL275.}
  \label{Fig:dpt_vis}
\end{figure}

Apart from the single RGB image and category-level shape prior, we also input additional geometric information into our pipeline to help build 2D-3D correspondences. Therefore, we conduct an ablation study on different formats of geometric information to justify the design choice of our method. Here, we mainly test on three representations, i.e. 3D point cloud, depth map, and normal map. In Figure \ref{Fig:dpt_vis}, we show some examples of generated 2.5D sketches for all 6 categories on both CAMERA25 and REAL275. These 2.5D geometric cues can roughly restore the instance-level information according to the input image patch. The point cloud used for evaluation is then back-projected from the (relative) depth map using intrinsics provided by NOCS datasets, and we further normalize it into a unit sphere to ease the difficulty of training. We also change the feature extractor to PointNet++~\cite{qi2017pointnet++} to cope with unordered point sets.

We report the results in Table \ref{Tab:ablation_input}. Despite point clouds directly coming from predicted depth maps, they have quite different performances in terms of the $10^\circ$ metric, where the depth-only setting is 3.2\% and 12.9\% higher on CAMERA25 and REAL275, respectively. And this is probably caused by the sparse sampling of depth maps. When dealing with 2.5D representations, the large receptive fields of CNN can enable certain robustness to noise, thus extracting more distinctive features. While the sampling of depth maps, instead, magnifies the impact of outliers due to the sparsity of point clouds. Furthermore, we discover that the combination of depth and normal maps has a positive influence on correspondence learning, resulting in 2.2\% and 2.7\% improvements on the $10^\circ$ metric on CAMERA25 and REAL275, respectively. This eventually validates our design choice of input modalities for additional geometric information.

\subsubsection{Effect of Scale Recovery} \label{Sec:exp_scale}

To evaluate the effectiveness of our proposed metric scale recovery strategy, we report the results with and without the usage of relative scale offset in Table \ref{Tab:ablation_scale}. It is well-known that recovering the metric scale from a single image is a highly ill-posed problem due to the ignorance of depth information. However, we manage to restore this value by leveraging the category-level statistics (i.e. mean scale) collected from datasets. The first row shows a particular case where only the mean scale of each category is used to object pose estimation. It already starts giving promising results compared to previous RGB-based methods, yet fine-tuning with relative scale offset (see the second row) can further boost the performance. We can observe consistent improvements on CAMERA25 while the performance slightly degrades on REAL275 regarding IoU50 and 10cm metrics. 

\begin{figure}[htbp]
  \centering
  \includegraphics[width=0.85\linewidth]{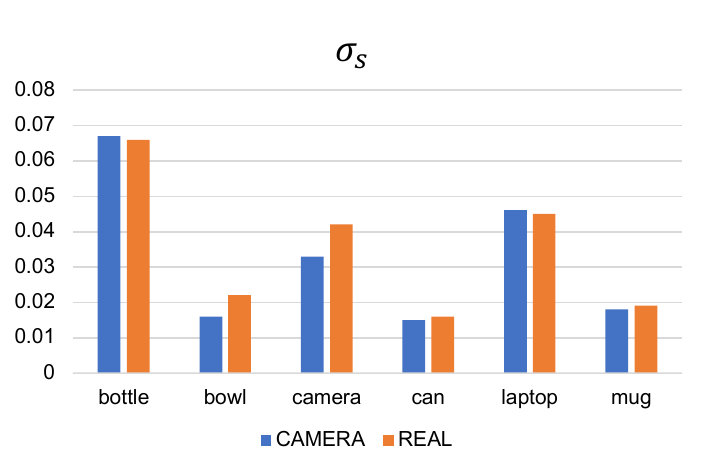}
  \caption{The standard deviation of ground truth metric scales w.r.t. the mean scale of each category on both CAMERA and REAL datasets.}
  \label{Fig:scale_std}
\end{figure}

To better understand the possible reasons behind this phenomenon, we calculate the standard deviation of ground truth metric scales w.r.t. the mean scale of each category $$\sigma_s = \sqrt{\frac{\sum (s_{gt} - s_r)^2}{k}}$$ where $k$ denotes the number of instances within each category. This is illustrated in Figure \ref{Fig:scale_std}. It clearly shows that synthetic objects and real objects in both datasets have roughly similar standard deviations, except for the bowl and camera classes. As for these two categories, the variation around the mean scale on the REAL dataset is higher than the CAMERA dataset. Besides, there is not enough training data in the REAL dataset so the network training is dominated by synthetic data. All of these factors combined may account for the degradation. And last but not least, the bottle, camera, and laptop classes have top3 highest standard deviations due to the large intra-class dissimilarity. This partially explains why they are inferior to other categories in category-level pose estimation.

\subsection{Limitations}

Through extensive experiments on the NOCS benchmark, we have witnessed a discrepancy between synthetic and real datasets. And this may be caused by several reasons. First, the quality of predicted 2.5D sketches is generally better for synthetic objects than real objects. Take the predictions in Figure \ref{Fig:dpt_vis} for example. The generated depth maps and normal maps appear blurrier on the edge of real objects due to the shadows and cluttered backgrounds. Second, real-world instances tend to have larger shape variations as indicated in Figure \ref{Fig:scale_std}, and certain categories need to be taken care of to deal with the inherent intra-class dissimilarity in category-level object pose estimation. Therefore, our future work will focus on compensating for the domain gap between synthetic and real data.

\section{CONCLUSIONS}

In this paper, we have presented a decoupled approach for RGB-based category-level object pose and size estimation, where 6D rigid transformation and metric scale are computed in a separate way. This strategy benefits the 6D pose estimation a lot as we explicitly stop the error propagation resulting from imperfect scale predictions. We leverage rich geometric information to build reliable 2D-3D correspondences, and a direct scale regression branch is used to recover the metric scale. Through extensive experiments, our proposed method has demonstrated significant improvements over existing SOTA RGB-based approaches.



\bibliographystyle{IEEEtran}
\bibliography{root}

\begin{thebibliography}{10}
\providecommand{\url}[1]{#1}
\csname url@samestyle\endcsname
\providecommand{\newblock}{\relax}
\providecommand{\bibinfo}[2]{#2}
\providecommand{\BIBentrySTDinterwordspacing}{\spaceskip=0pt\relax}
\providecommand{\BIBentryALTinterwordstretchfactor}{4}
\providecommand{\BIBentryALTinterwordspacing}{\spaceskip=\fontdimen2\font plus
\BIBentryALTinterwordstretchfactor\fontdimen3\font minus
  \fontdimen4\font\relax}
\providecommand{\BIBforeignlanguage}[2]{{%
\expandafter\ifx\csname l@#1\endcsname\relax
\typeout{** WARNING: IEEEtran.bst: No hyphenation pattern has been}%
\typeout{** loaded for the language `#1'. Using the pattern for}%
\typeout{** the default language instead.}%
\else
\language=\csname l@#1\endcsname
\fi
#2}}
\providecommand{\BIBdecl}{\relax}
\BIBdecl

\bibitem{tan20176d}
D.~J. Tan, N.~Navab, and F.~Tombari, ``6d object pose estimation with depth
  images: A seamless approach for robotic interaction and augmented reality,''
  \emph{arXiv preprint arXiv:1709.01459}, 2017.

\bibitem{su2019deep}
Y.~Su, J.~Rambach, N.~Minaskan, P.~Lesur, A.~Pagani, and D.~Stricker, ``Deep
  multi-state object pose estimation for augmented reality assembly,'' in
  \emph{2019 IEEE International Symposium on Mixed and Augmented Reality
  Adjunct (ISMAR-Adjunct)}.\hskip 1em plus 0.5em minus 0.4em\relax IEEE, 2019,
  pp. 222--227.

\bibitem{du2021vision}
G.~Du, K.~Wang, S.~Lian, and K.~Zhao, ``Vision-based robotic grasping from
  object localization, object pose estimation to grasp estimation for parallel
  grippers: a review,'' \emph{Artificial Intelligence Review}, vol.~54, no.~3,
  pp. 1677--1734, 2021.

\bibitem{wang2020feature}
C.~Wang, X.~Zhang, X.~Zang, Y.~Liu, G.~Ding, W.~Yin, and J.~Zhao, ``Feature
  sensing and robotic grasping of objects with uncertain information: A
  review,'' \emph{Sensors}, vol.~20, no.~13, p. 3707, 2020.

\bibitem{xu2022rnnpose}
Y.~Xu, K.-Y. Lin, G.~Zhang, X.~Wang, and H.~Li, ``Rnnpose: Recurrent 6-dof
  object pose refinement with robust correspondence field estimation and pose
  optimization,'' in \emph{Proceedings of the IEEE/CVF Conference on Computer
  Vision and Pattern Recognition}, 2022, pp. 14\,880--14\,890.

\bibitem{iwase2021repose}
S.~Iwase, X.~Liu, R.~Khirodkar, R.~Yokota, and K.~M. Kitani, ``Repose: Fast 6d
  object pose refinement via deep texture rendering,'' in \emph{Proceedings of
  the IEEE/CVF International Conference on Computer Vision}, 2021, pp.
  3303--3312.

\bibitem{lin2022category}
J.~Lin, Z.~Wei, C.~Ding, and K.~Jia, ``Category-level 6d object pose and size
  estimation using self-supervised deep prior deformation networks,'' in
  \emph{Computer Vision--ECCV 2022: 17th European Conference, Tel Aviv, Israel,
  October 23--27, 2022, Proceedings, Part IX}.\hskip 1em plus 0.5em minus
  0.4em\relax Springer, 2022, pp. 19--34.

\bibitem{chen2021sgpa}
K.~Chen and Q.~Dou, ``{SGPA: Structure-guided prior adaptation for
  category-level 6D object pose estimation},'' in \emph{{Proceedings of the
  IEEE/CVF International Conference on Computer Vision}}, 2021, pp. 2773--2782.

\bibitem{lin2022sar}
H.~Lin, Z.~Liu, C.~Cheang, Y.~Fu, G.~Guo, and X.~Xue, ``Sar-net: shape
  alignment and recovery network for category-level 6d object pose and size
  estimation,'' in \emph{Proceedings of the IEEE/CVF Conference on Computer
  Vision and Pattern Recognition}, 2022, pp. 6707--6717.

\bibitem{nocs}
H.~Wang, S.~Sridhar, J.~Huang, J.~Valentin, S.~Song, and L.~J. Guibas,
  ``{Normalized object coordinate space for category-level 6D object pose and
  size estimation},'' in \emph{{Proceedings of the IEEE/CVF Conference on
  Computer Vision and Pattern Recognition}}, 2019, pp. 2642--2651.

\bibitem{di2022gpv}
Y.~Di, R.~Zhang, Z.~Lou, F.~Manhardt, X.~Ji, N.~Navab, and F.~Tombari,
  ``Gpv-pose: Category-level object pose estimation via geometry-guided
  point-wise voting,'' in \emph{Proceedings of the IEEE/CVF Conference on
  Computer Vision and Pattern Recognition}, 2022, pp. 6781--6791.

\bibitem{lin2021dualposenet}
J.~Lin, Z.~Wei, Z.~Li, S.~Xu, K.~Jia, and Y.~Li, ``Dualposenet: Category-level
  6d object pose and size estimation using dual pose network with refined
  learning of pose consistency,'' in \emph{Proceedings of the IEEE/CVF
  International Conference on Computer Vision}, 2021, pp. 3560--3569.

\bibitem{chen2021fs}
W.~Chen, X.~Jia, H.~J. Chang, J.~Duan, L.~Shen, and A.~Leonardis, ``Fs-net:
  Fast shape-based network for category-level 6d object pose estimation with
  decoupled rotation mechanism,'' in \emph{Proceedings of the IEEE/CVF
  Conference on Computer Vision and Pattern Recognition}, 2021, pp. 1581--1590.

\bibitem{chen2020synthesis}
X.~Chen, Z.~Dong, J.~Song, A.~Geiger, and O.~Hilliges, ``{Category level object
  pose estimation via neural analysis-by-synthesis},'' in \emph{{Proceedings of
  the European Conference on Computer Vision}}.\hskip 1em plus 0.5em minus
  0.4em\relax Springer, 2020, pp. 139--156.

\bibitem{fan2022depth}
Z.~Fan, Z.~Song, J.~Xu, Z.~Wang, K.~Wu, H.~Liu, and J.~He, ``{Object level
  depth reconstruction for category level 6D object pose estimation from
  monocular RGB image},'' in \emph{{Proceedings of the European Conference on
  Computer Vision}}.\hskip 1em plus 0.5em minus 0.4em\relax Springer, 2022, pp.
  220--236.

\bibitem{lee2021shape}
T.~Lee, B.-U. Lee, M.~Kim, and I.~S. Kweon, ``{Category-level metric scale
  object shape and pose estimation},'' \emph{{IEEE Robotics and Automation
  Letters}}, vol.~6, no.~4, pp. 8575--8582, 2021.

\bibitem{eftekhar2021omnidata}
A.~Eftekhar, A.~Sax, J.~Malik, and A.~Zamir, ``{Omnidata: A scalable pipeline
  for making multi-task mid-level vision datasets from 3d scans},'' in
  \emph{{Proceedings of the IEEE/CVF International Conference on Computer
  Vision}}, 2021, pp. 10\,786--10\,796.

\bibitem{tian2020shape-prior}
M.~Tian, M.~H. Ang, and G.~H. Lee, ``{Shape prior deformation for categorical
  6D object pose and size estimation},'' in \emph{{Proceedings of the European
  Conference on Computer Vision}}.\hskip 1em plus 0.5em minus 0.4em\relax
  Springer, 2020, pp. 530--546.

\bibitem{bruns2023rgbdbased}
L.~Bruns and P.~Jensfelt, ``{RGB-D-Based Categorical Object Pose and Shape
  Estimation: Methods, Datasets, and Evaluation},'' 2023.

\bibitem{he2017maskrcnn}
K.~He, G.~Gkioxari, P.~Doll{\'a}r, and R.~Girshick, ``{Mask R-CNN},'' in
  \emph{{Proceedings of the IEEE International Conference on Computer Vision}},
  2017, pp. 2961--2969.

\bibitem{umeyama1991umeyama}
S.~Umeyama, ``Least-squares estimation of transformation parameters between two
  point patterns,'' \emph{{IEEE Transactions on Pattern Analysis \& Machine
  Intelligence}}, vol.~13, no.~04, pp. 376--380, 1991.

\bibitem{chen2020learning}
D.~Chen, J.~Li, Z.~Wang, and K.~Xu, ``{Learning canonical shape space for
  category-level 6D object pose and size estimation},'' in \emph{{Proceedings
  of the IEEE/CVF Conference on Computer Vision and Pattern Recognition}},
  2020, pp. 11\,973--11\,982.

\bibitem{wang2021cascaded}
J.~Wang, K.~Chen, and Q.~Dou, ``{Category-level 6D object pose estimation via
  cascaded relation and recurrent reconstruction networks},'' in
  \emph{{Proceedings of the IEEE/RSJ International Conference on Intelligent
  Robots and Systems}}, 2021, pp. 4807--4814.

\bibitem{lin2022selfsupervised}
J.~Lin, Z.~Wei, C.~Ding, and K.~Jia, ``{Category-level 6D object pose and size
  estimation using self-supervised deep prior deformation networks},'' in
  \emph{{Proceedings of the European Conference on Computer Vision}}.\hskip 1em
  plus 0.5em minus 0.4em\relax Springer, 2022, pp. 19--34.

\bibitem{manhardt2020cps}
F.~Manhardt, G.~Wang, B.~Busam, M.~Nickel, S.~Meier, L.~Minciullo, X.~Ji, and
  N.~Navab, ``{CPS++: Improving Class-level 6D Pose and Shape Estimation From
  Monocular Images With Self-Supervised Learning},'' 2020.

\bibitem{Marr1982}
D.~Marr, \emph{Vision}.\hskip 1em plus 0.5em minus 0.4em\relax W. H. Freeman,
  1982.

\bibitem{ranftl2021dpt}
R.~Ranftl, A.~Bochkovskiy, and V.~Koltun, ``Vision transformers for dense
  prediction,'' in \emph{{Proceedings of the IEEE/CVF International Conference
  on Computer Vision}}, 2021, pp. 12\,179--12\,188.

\bibitem{kar2022augmentation}
O.~F. Kar, T.~Yeo, A.~Atanov, and A.~Zamir, ``{3D Common Corruptions and Data
  Augmentation},'' in \emph{{Proceedings of the IEEE/CVF Conference on Computer
  Vision and Pattern Recognition}}, 2022, pp. 18\,963--18\,974.

\bibitem{zhao2017psp}
H.~Zhao, J.~Shi, X.~Qi, X.~Wang, and J.~Jia, ``Pyramid scene parsing network,''
  in \emph{{Proceedings of the IEEE/CVF Conference on Computer Vision and
  Pattern Recognition}}, 2017, pp. 2881--2890.

\bibitem{he2016resnet}
K.~He, X.~Zhang, S.~Ren, and J.~Sun, ``{Deep residual learning for image
  recognition},'' in \emph{{Proceedings of the IEEE conference on computer
  vision and pattern recognition}}, 2016, pp. 770--778.

\bibitem{qi2017pointnet++}
C.~R. Qi, L.~Yi, H.~Su, and L.~J. Guibas, ``{Pointnet++: Deep hierarchical
  feature learning on point sets in a metric space},'' \emph{{Advances in
  neural information processing systems}}, vol.~30, 2017.

\end{thebibliography}

\end{document}